%% file: KPCA_Reg_ECCV.tex
\documentclass[runningheads]{llncs}
\usepackage{graphicx}
\usepackage{amsmath,amssymb} 
\usepackage{color}
\usepackage[width=122mm,left=12mm,paperwidth=146mm,height=193mm,top=12mm,paperheight=217mm]{geometry}

\usepackage{epsfig}
\usepackage{graphicx}
\usepackage{array}

\usepackage{url}
\usepackage{color}
\usepackage{multirow}

\usepackage[norelsize,ruled,vlined]{algorithm2e}
\usepackage{wrapfig}
\usepackage{wraptable}
\usepackage[font=small,labelfont=bf]{caption}

\include{Macros}

\begin{document}
\pagestyle{headings}
\mainmatter
\title{Non-linear Dimensionality Regularizer for Solving Inverse Problems}

\titlerunning{Non-linear Dimensionality Regularizer for Solving Inverse Problems}

\authorrunning{Garg et. al.}

\author{Ravi Garg \and  Anders Eriksson \and Ian Reid}
\institute{University of Adelaide \and Queensland University of Technology \and University of Adelaide}

\maketitle

\begin{abstract}

 Consider an ill-posed inverse problem of estimating causal factors from observations, 
 one of which is known to lie near some (unknown) 
   low-dimensional, non-linear manifold expressed by a predefined Mercer-kernel.
   Solving this problem requires simultaneous estimation of these factors
   and learning the low-dimensional representation for them.
   In this work, we introduce a novel  non-linear dimensionality regularization technique for solving such problems without pre-training. 
%
   
   We re-formulate Kernel-PCA as an energy minimization problem in which low dimensionality constraints are introduced as regularization terms in the energy.  To the best of our knowledge, ours is the first attempt to
   create a dimensionality regularizer in the KPCA framework.
   Our approach relies on robustly penalizing the rank of the recovered factors
   directly in the implicit feature space 
   to create their low-dimensional approximations in closed form.
   
   
   
 Our approach performs robust KPCA in the presence of missing data and noise.  We demonstrate  state-of-the-art results on predicting missing entries in the standard oil flow dataset.
   Additionally, we evaluate our method on the challenging problem of Non-Rigid Structure from Motion and our approach
   delivers promising
   results on CMU mocap dataset despite the presence of significant occlusions and noise.
    
\end{abstract}

Dimensionality reduction techniques are widely used in data modeling, visualization
and unsupervised learning. Principal component analysis (PCA\cite{jolliffe2002principal}), Kernel PCA (KPCA\cite{scholkopf1998nonlinear}) and Latent Variable
Models (LVMs\cite{lawrence2005probabilistic}) are some of the well known techniques used to create low dimensional representations
of the given data while preserving its significant information.

\begin{figure}[!t]
\caption{\textbf{Non-linear dimensionality regularisation improves NRSfM performance compared to its linear counterpart.}
   Figure shows the ground truth 3D structures in red wire-frame
   overlaid with the structures estimated using:
   (a) proposed non-linear dimensionality regularizer shown in blue dots and  
   (b) corresponding linear dimensionality regularizer (TNH) shown in black crosses, for sample frames of CMU mocap sequence.
   Red circles represent the 3D points for which the projections were known whereas squares annotated missing 2D observations.
   See text and Table \ref{tab:NRSfM_GTrot} for details.
   \label{fig:results_nrsfm}}
  \centering
   \includegraphics[width=.85\columnwidth]{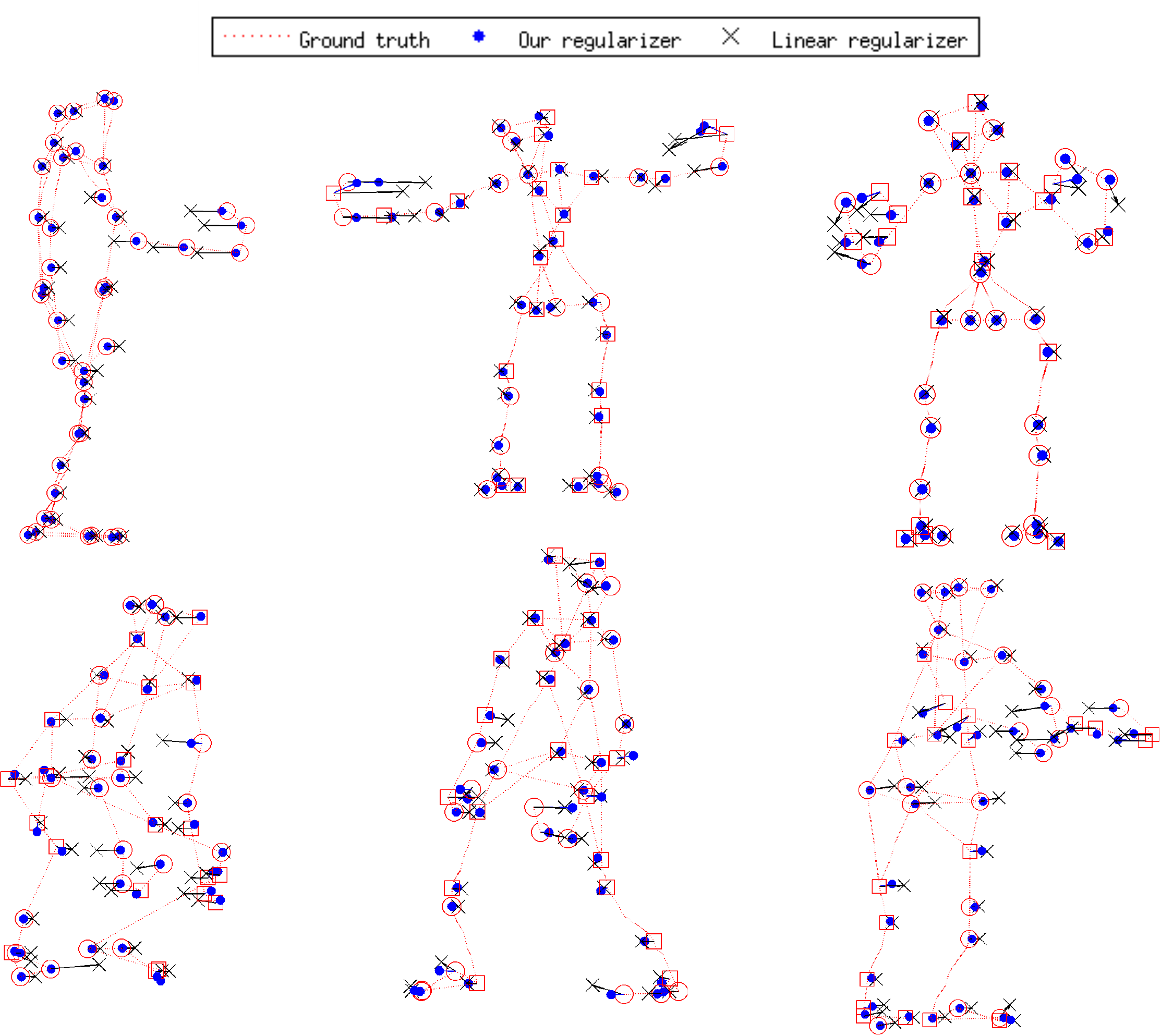}
   \vspace{-5mm}
\end{figure}

One key deployment of low-dimensional modeling occurs in solving ill-posed inference problems.
Assuming the valid solutions to the problem lie near a low-dimensional manifold (i.e. can be parametrized with a 
reduced set of variables) allows for a tractable inference for otherwise under-constrained problems.
After the seminal work of \cite{candes2009exact,recht2010guaranteed} on guaranteed rank minimization 
 of the matrix via trace norm heuristics \cite{fazel2002matrix},
many ill-posed computer vision problems have been tackled by using the trace norm --- a convex surrogate of the rank function ---
as a regularization term in an energy minimization framework\cite{candes2009exact,zhou2014low}. 
The flexible and easy integration of low-rank priors is one of key factors for versatility and success of many algorithms.
For example, pre-trained active appearance models \cite{cootes2001active} or 3D morphable models \cite{blanz1999morphable}
are converted to robust feature tracking \cite{poling2014better}, dense registration \cite{garg2013variational} 
and vivid reconstructions of natural videos \cite{garg2013dense} with no {\em a priori} knowledge of the scene. 
Various bilinear factorization problems like background modeling, structure from motion or photometric stereo are also addressed
with a variational formulation of the trace norm regularization \cite{CabralDCB13}.

On the other hand, although many non-linear dimensionality reduction techniques --- in particular KPCA --- have been shown to outperform their
linear counterparts for many data modeling tasks, they are seldom used to solve inverse problems without using a training phase.  
A general (discriminative) framework for using non-linear dimensionality reduction is: (i) learn a low-dimensional
representation for the data using training examples via the kernel trick (ii) project the test examples on the learned manifold
and finally (iii) find a data point (pre-image) corresponding to each projection in the input space. 

This setup has two major disadvantages. Firstly, many problems of interest come with corrupted observations --- noise,
missing data and outliers --- which violate the low-dimensional modeling assumption.
Secondly, computing the pre-image of any point in the
low dimensional feature subspace is non-trivial: the pre-image for many points in the low dimensional space might not even exist because 
the non linear feature mapping function used for mapping the data from input space to the feature space is non-surjective.
%

\begin{figure}[h]
\centering
    \includegraphics[width=.9\textwidth]{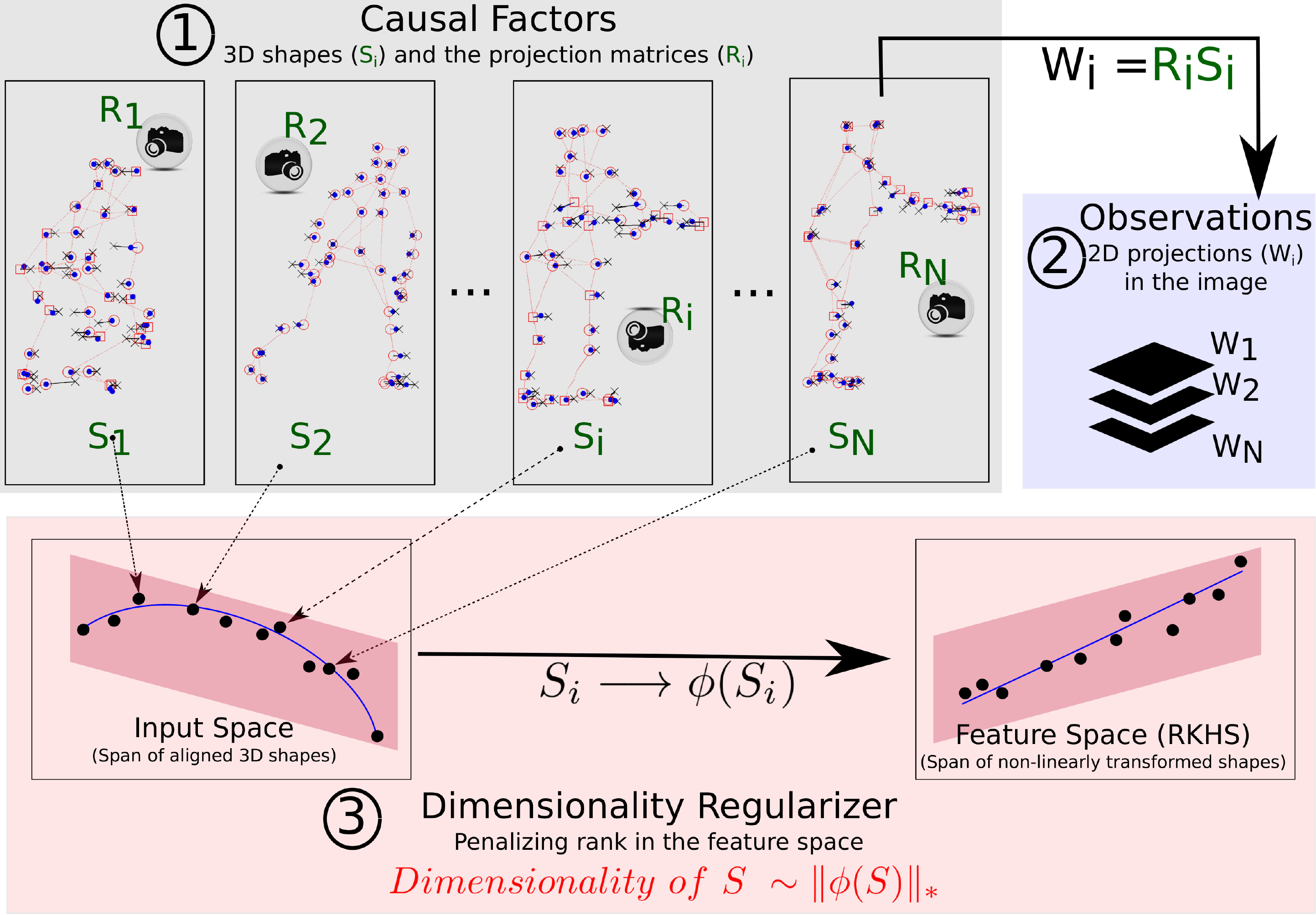}
   \caption{\label{fig:intro_nrsfm}\textbf{Non-linear dimensionality regularizer for NRSfM.} The top part of the figure explains the ill-posed \textit{inverse problem} of recovering the \textit{causal factors (1)}; projection matrices $R_i$ and 3D structures $S_i$, 
   from 2D image \textit{observations (2)} $W_i$'s,  by minimizing the \textit{image reprojection} error  $f(W,R,S) = \sum_i \|W_i-R_iS_i\|^2$.
   Assuming that the recovered 3D structures ($S_i$'s) lies near an unknown non-linear manifold (represented by the blue curve) in the input space, 
   we propose to \textit{regularize the dimensionality of this manifold (3)} --- span of the non-linearly transformed shape vectors $\phi(S_i)$'s  --- by minimizing $\| \phi(S) \|_{*}$. 
   The non-linear transformation $\phi$ is defined implicitly with a Mercer kernel and maps the non-linear manifold to a linear low rank subspace (shown in blue line) of RKHS.}%
   \vspace{-5mm}
\end{figure}

Previously, extensions to KPCA like Robust KPCA (RKPCA\cite{nguyen2009robust}) and probabilistic KPCA 
(PKPCA\cite{sanguinetti2006missing}) with missing data have been proposed to address the first concern, while
various additional regularizers have been used to estimate the pre-image robustly \cite{bakir2004learning,mika1998kernel,kwok2004pre,abrahamsen2009input}. 

Generative models like LVMs \cite{lawrence2005probabilistic} are often used for inference
by searching the low-dimensional latent space for a location 
which maximizes the likelihood of the 
observations. Problems like segmentation, tracking and semantic 3D reconstruction \cite{prisacariu2011nonlinear,dame2013dense} greatly benefit from using LVM. 
However, the latent space is learned \emph{a priori} with clean training data in all these approaches.

Almost all non-linear dimensionality reduction techniques are
non-trivial to generalize for solving ill-posed problems (See section \ref{sec:KNRSfM}) without a pre-training stage.
Badly under-constrained problems require the low-dimensional constraints even for finding an initial solution,
eliminating applicability of the standard ``projection + pre-image estimation'' paradigm.   
This hinders the utility of non-linear dimensionality reduction
and a suitable regularization technique to penalize the non-linear dimensionality is desirable.

\textbf{Sum and Substance:} A closer look at most non-linear dimensionality reduction techniques reveals that they rely upon a non-linear mapping function
which maps the data from input space to a (usually) higher dimensional feature space. 
In this feature space the data is assumed to lie on a low-dimensional hyperplane --- thus, \textit{linear low-rank prior is apt in the feature space}.
Armed with this simple observation, our aim is to focus on incorporating the advances made in linear  
dimensionality reduction techniques 
to their non-linear counterparts,  
while addressing the problems described above.
Figure \ref{fig:intro_nrsfm} explains this central idea and proposed dimensionality regularizer in a nutshell with Non Rigid Structure from Motion (NRSfM) as the example application.

\textbf{Our Contribution:} In this work we propose a unified for simultaneous robust KPCA and pre-image estimation while solving
an ill-posed inference problem without a pre-training stage.


In particular we propose a novel robust energy minimization algorithm which handles the implicitness of the feature space to directly
penalize its rank by iteratively:
\begin{itemize}\vspace{-1mm}
 \item creating robust low-dimensional representation for the data given the
kernel matrix in closed form and\vspace{-1mm}
\item reconstructing the noise-free version of the data (pre-image of the features space projections) 
using the estimated low-dimensional representations in a unified framework.
\end{itemize}\vspace{-1mm}

The proposed algorithm: (i) provides a novel closed form solution to robust KPCA; 
(ii) yields state-of-the-art results on missing data prediction for the well-known oil flow dataset; 
(iii) outperforms state-of-the-art linear dimensionality (rank) regularizers to solve NRSfM;
and (iv) can be trivially generalized to incorporate other cost functions in an energy minimization framework to solve various ill-posed inference problems.
%
%
%
%
%
%
%
%
%
%

\section{Problem formulation}\label{sec:Formulation}
This paper focuses on solving a generic inverse problem of recovering causal factor $S = [s_1, \ s_2, \ \cdots \, s_N ] \in \mathcal{X} \times N$
from $N$ observations $W = [w_1, \ w_2, \ \cdots \, w_N ] \in \mathcal{Y} \times N$ such that $f(W,S) = 0$. 
Here function $f(\textnormal{observation,variable})$, is a generic loss function which aligns the observations $W$ with the variable $S$ (possibly
via other causal factors. e.g. $R$ or $Z$ in Section \ref{sec:MatrixComp} and \ref{sec:KNRSfM}).

If, $f(W,S)=0$ 
is ill-conditioned (for example when $\mathcal{Y} \ll \mathcal{X}$), 
we want to recover matrix $S$ under the assumption that the columns of it lie near a low-dimensional non-linear manifold. 
This can be done by solving a constrained optimization problem of the following form:
\begin{eqnarray}
 \min_{S}  &rank(\Phi (S)) \nonumber \\
 & s.t. \ \ f(W,S) \leq \epsilon \label{eq:dimRed}
\end{eqnarray}
where $\Phi(S) = [\phi(s_1), \ \phi(s_2), \ \cdots, \ \phi(s_N)] \in \mathcal{H} \times N$
is the non-linear mapping of matrix $S$ from the input space $\mathcal{X}$ to the feature space $\mathcal{H}$
(also commonly referred as Reproducing Kernel Hilbert Space), via a non-linear mapping function
$\phi: \mathcal{X} \to \mathcal{H}$ associated with a Mercer kernel $K$ such that $K(S)_{i,j} = \phi(s_i)^T\phi(s_j)$. 

In this paper we present a novel energy minimization framework to solve problems of the general form \eqref{eq:dimRed}.



As our first contribution, we relax the problem \eqref{eq:dimRed} by using the trace norm of $\Phi(S)$ --- the convex surrogate of rank function ---
as a penalization function. The trace norm $\| M \|_{*} =: \sum_i \lambda_i(M)$ of a matrix $M$ is the sum of its eigenvalues $\lambda_i(M)$ and was 
proposed as a tight convex relaxation\footnote{More precisely, $\| M \|_{*}$ was shown to be the tight convex envelope of $rank(M)/\|M\|_{s}$, where $\|M\|_s$ represent spectral norm of $M$.}
of the $rank(M)$ and is used in many vision problems as a rank regularizer \cite{fazel2002matrix}.
Although the rank minimization via trace norm relaxation does not lead to a convex problem
in presence of a non-linear kernel function, we show in \ref{sec:CFS} that it leads to a closed-form
solution to denoising a kernel matrix via penalizing the rank of recovered data ($S$) directly in the feature space. 

With these changes we can rewrite  \eqref{eq:dimRed} as:
\begin{equation}
 \min_{S} \ \ f(W,S) + \tau \| \Phi (S) \|_{*}  \label{eq:objectGen}
\end{equation}
where $\tau$ is a regularization strength.\footnote{$1/\tau$ can also be viewed as Lagrange multiplier to the constraints in \eqref{eq:dimRed}.}

It is important to notice that although the rank of the kernel matrix $K(S)$ is equal to the rank of $\Phi(S)$, $\|K(S)\|_*$ is merely $\|\Phi(S)\|_\mathcal{F}^2$.
Thus, directly penalizing 
the sum of the singular values of $K(S)$ will not encourage 
low-rank in the feature space.%
%
\footnote{Although it is clear that relaxing the rank of kernel matrix to $\|K(S)\|_{*}$ is suboptimal, works like \cite{HuangSD12,CabralDCB13} with a variational definition of nuclear norm, allude to the possibility of kernelization.  Further investigation is required to compare this counterpart to our tighter relaxation.}


Although we have relaxed the non-convex rank function, \eqref{eq:objectGen} is in general difficult 
to 
minimize due to the implicitness of the feature space. Most widely used kernel functions like
RBF do not have a explicit definition of the function $\phi$. Moreover, the feature space
for many kernels is high- (possibly infinite-) dimensional, leading to intractability.
These issues are identified as the main barriers to
robust KPCA and pre-image estimation \cite{nguyen2009robust}.
Thus, we have to reformulate \eqref{eq:objectGen} by applying kernel trick where 
the cost function \eqref{eq:objectGen} can be expressed in terms of the kernel function alone.

The key insight here is that under the assumption that kernel matrix $K(S)$ is positive semidefinite,
we can factorize it as: $K(S) = C^TC$. Although, this factorization is non-unique, it is trivial to
show the following: 
\begin{eqnarray}
 &\sqrt{\lambda_i(K(S))} = \lambda_i(C) = \lambda_i(\Phi(S))  \nonumber\\
 \text{Thus:} &\|C\|_{*} =  \|\Phi(S)\|_{*}  \ \  \ \forall \ C  : C^T C = K(S) \label{eq:proof_Relexation} 
\end{eqnarray}
where $\lambda_i(.)$ is the function mapping the input matrix to its $i^{th}$ largest eigenvalue.

The row space of matrix $C$ in \eqref{eq:proof_Relexation} can be seen to span the eigenvectors associated 
with the kernel matrix $K(S)$ --- hence the principal components of the non-linear manifold we want to estimate.

Using \eqref{eq:proof_Relexation}, problem \eqref{eq:objectGen} can finally be written as:
\begin{align}
 \min_{S,C}&  \ \ f(W,S) + \tau \| C \|_{*} \nonumber \\
 &s. t. \ \ \   K(S) = C^T C \label{eq:HardConstraint}
\end{align}
The above minimization can be solved with a soft relaxation of the manifold constraint
by assuming that the columns of $S$ lie near the non-linear manifold.
\begin{eqnarray}
 \min_{S,C} \ \  f(W,S) +  \frac{\rho}{2} \|K(S) - C^T C\|^2_\mathcal{F} + \tau \| C \|_{*}  \label{eq:final_energy} 
\end{eqnarray}
As $\rho \rightarrow \infty$, the optimum of \eqref{eq:final_energy} approaches the optimum of \eqref{eq:HardConstraint} .
A local optimum of \eqref{eq:HardConstraint} can be achieved using the penalty method of \cite{Nocedal}
by optimizing \eqref{eq:final_energy} while iteratively increasing  $\rho$ as explained in Section \ref{sec:Optimization}.

Before moving on, we would like to discuss some alternative interpretations of \eqref{eq:final_energy} 
and its relationship to previous work -- in particular LVMs.
Intuitively, we can also interpret \eqref{eq:final_energy} from the probabilistic viewpoint as commonly used in
latent variable model based approaches to define kernel function \cite{lawrence2005probabilistic}.
For example a RBF kernel with additive Gaussian noise and inverse width $\gamma$ can be defined as:
$K(S)_{i,j} = e^{-\gamma \|s_i - s_j\|^2} + \epsilon$, where $\epsilon \sim \mathcal{N}(0,\sigma)$. 
In other words, with a finite $\rho$, our \textit{model allows the data points to lie near a non-linear 
low-rank manifold} instead of on it.
Its worth noting here that like LVMs, our energy formulation also attempts to maximize the likelihood
of regenerating the training data $W$, (by choosing $f(W,S)$ to be a simple least squares cost) while doing dimensionality reduction.

Note that in closely related work \cite{geiger2009rank}, continuous rank penalization (with a logarithmic prior)
has also been used for robust probabilistic non-linear dimensionality reduction and model selection in LVM framework. 
However, unlike \cite{geiger2009rank,lawrence2005probabilistic} where the non-linearities are modeled in latent space (of predefined dimensionality), our 
approach directly penalizes the non-linear dimensionality of data in a KPCA framework and is applicable to solve inverse problems without pre-training.

\section{Optimization}\label{sec:Optimization}
We approach the optimization of \eqref{eq:final_energy} by solving
the following two sub-problems in alternation:
\begin{align} 
  \min_{S}  \ \ \  f(W,S) +  \frac{\rho}{2} \|K(S) - C^T C\|^2_\mathcal{F} \label{eq:E_inv} \\
  \min_{C}  \ \ \  \tau \| C \|_{*} + \frac{\rho}{2} \|K(S) - C^T C\|^2_\mathcal{F}  \label{eq:E_Dreduc}
\end{align}
Algorithm \ref{alg:altern} outlines the approach and we give a detailed description and interpretations of both sub-problems \eqref{eq:E_Dreduc} and \eqref{eq:E_inv} in next two sections of the paper.
%
\SetAlFnt{\footnotesize}\SetKwInput{KwData}{Parameters}
\begin{algorithm}
\KwIn{Initial estimate $S^0$ of $S$.}
\KwOut{Low-dimensional $S$ and kernel representation $C$.}
\KwData{Initial $\rho^0$ and maximum $\rho_{max}$ penalty, with scale $\rho_s$.}
\BlankLine
- $S = S^0,\rho = \rho^0$ \; 

\While{$\rho \leq \rho_{max}$}{
\While{not converged}{
 - Fix $S$ and estimate $C$ via closed-form solution of \eqref{eq:E_Dreduc} using Algorithm \ref{alg:CFS}\;
 - Fix $C$ and minimize \eqref{eq:E_inv} to update $S$ using LM algorithm\;
 
}- $\rho =  \rho \rho_s$ \;}
\caption{Inference with Proposed Regularizer.}\label{alg:altern}
\end{algorithm}\vspace{-3mm}


\subsection{Pre-image estimation to solve inverse problem.}\label{sec:Pre-image}
Subproblem \eqref{eq:E_inv} can be seen as a generalized pre-image estimation problem:  we seek the factor $s_i$, which is the pre-image of the projection of $\phi(s_i)$ onto the principle subspace of the RKHS stored in $C^TC$, which best explains the observation $w_i$. 
Here \eqref{eq:E_inv} is generally a non-convex problem, unless the
Mercer-kernel is linear, and must therefore be solved using non-linear optimization techniques.
In this work, we use the Levenberg-Marquardt algorithm for optimizing \eqref{eq:E_inv}.

Notice that \eqref{eq:E_inv} only computes the pre-image for the feature space projections of the data points
with which the non-linear manifold
(matrix $C$) is learned. An extension to our formulation is desirable if one wants to use the learned
non-linear manifold for denoising test data in a classic pre-image estimation framework. Although 
a valuable direction to pursue, it is out of scope of the present paper.

%
%

\subsection{Robust dimensionality reduction}\label{sec:CFS}
One can interpret sub-problem \eqref{eq:E_Dreduc} as a robust form of KPCA where the kernel
matrix has been corrupted with Gaussian noise and we want to generate its low-rank approximation.
Although \eqref{eq:E_Dreduc} is non-convex we can solve it in closed-form via singular value decomposition.
Our closed-form solution is outlined in Algorithm \ref{alg:CFS} and is based on the following theorem:

%
%
%
 \begin{theorem}\label{th:CFS}
 With $\S^n \ni A \succeq 0$ let $A=U \Sigma U^T$ denote its singular value decomposition.  
 Then
 \begin{align}
 \min_L \ \ \frac{\rho}{2} \normF{A-L^T L} + \tau \normnuc{L} \hspace{10mm} \label{eq1a} \\
  =\sum_{i=1}^n \left(  \frac{\rho}{2} (\sigma_i- \gamma_i^{*2} )^2 + \tau \gamma_i^{*}  \right).
 \label{eq1b}
 \end{align}
 A minimizer $L^*$ of \eqref{eq1a} is given by 
 \begin{align}
 L^*=  \Gamma^* U^T 
 \label{eq1c}
 \end{align}
 with 
 $\Gamma^* \in \D_+^n$,
 $\gamma^*_i \in \{ \alpha \in \R_+ \ | \ p_{\sigma_i,\tau / 2\rho}(\alpha)=0 \} 
 \ \bigcup \ \{0\}$, where  $p_{a,b}$ denotes the depressed cubic 
 $ p_{a,b} (x) = x^3- a x + b $. 
$\D_+^n$ is the set of n-by-n diagonal matrices with non-negative entries. 
\end{theorem}

As the closed-form solution to \eqref{eq:E_Dreduc} is a key contribution of this work,
an extended proof to the above theorem is included in the Appendix \ref{Appendix:Proof}. 
Theorem \ref{th:CFS} shows that each eigenvalue of the minimizer $C^{*}$ of \eqref{eq:E_Dreduc} 
can be obtained by solving a depressed cubic whose coefficients are determined by the corresponding
eigenvalue of the kernel matrix and the regularization strength $\tau$.  The roots of each cubic, together with zero, comprise a set of candidates for the corresponding egienvalue of $C^*$.  The best one from this set is obtained by choosing the value which minimizes \eqref{eq1b} (see Algorithm \ref{alg:CFS}).
%


\SetAlFnt{\footnotesize}\SetKwInput{KwData}{Parameters}
\begin{algorithm}
\KwIn{Current estimate of $S$.}
\KwOut{Low-dimensional representation $C$.}
\KwData{Current $\rho$ and regularization strength $\tau$.}
\BlankLine\BlankLine
- $[U \ \Lambda \ U^T]$ = Singular Value Decomposition of $K(S)$\;
\tcp{$\Lambda$ is a diagonal matrix, storing $N$ singular values $\lambda_i$ of $K(S)$.}
\For{$i = 1$ \KwTo $N$}{
- Find three solutions ($l_r : r \in \{1, 2, 3\}$) of:  \\ \ \ \ \ \ \ \ \ \ \ \ \  \ \ \ \ \ \ \ \  \ \ \ \ \ 
  $l^3 - l \lambda_i +\dfrac {\tau}{2\rho} = 0$ \;
- set $l_4 = 0$\;
- $l_r = \max(l_r,0) \ \ \forall r \in \{ 1,2,3,4 \}$   \;
- $r = arg\min\limits_r \{{\frac{\rho}{2} \|\lambda_i - l_r^2\|^2 +  \tau \l_r}$ \} \;
- $\bar{\lambda}_i = l_r$ \; \
}
- $C = \bar{\Lambda} U^T$\;
\tcp{$\bar{\Lambda}$ is diagonal matrix storing $\bar{\lambda}_i$.}
\caption{Robust Dimensionality Reduction.}\label{alg:CFS}
\end{algorithm}

\section{Experiments}\label{sec:Experiments}
In this section we demonstrate the utility of the proposed algorithm.
The aims of our experiments are twofold: (i) to compare our dimensionality reduction technique
favorably with KPCA and its robust variants; and (ii) to demonstrate that the proposed non-linear dimensionality regularizer 
consistently outperforms its linear counterpart (a.k.a. nuclear norm) in solving inverse problems.

\subsection{Validating the closed form solution}
Given the relaxations proposed in Section \ref{sec:Formulation}, our assertion that the novel trace regularization
based non-linear dimensionality reduction is robust need to be substantiated.
To that end, we evaluate our closed-form solution of Algorithm \ref{alg:CFS} on the standard oil flow dataset introduced in \cite{bishop1993analysis}.

This dataset comprises 1000 training and 1000 testing data samples,
each of which is of 12 dimensions and categorized into one of three different classes. We add zero mean
Gaussian noise with variance $\sigma$ to the training data\footnote{Note that our formulation assumes Gaussian noise in $K(S)$ where as for this evaluation we add noise to $S$ directly.}
and recover the low-dimensional manifold for this noisy training data $S_\sigma$
with KPCA and contrast this with the results from  Algorithm \ref{alg:CFS}. An inverse width of the Gaussian kernel $\gamma = 0.075$ is used for all the experiments on the oil flow dataset.

It is important to note that in this experiment, we only estimate the principal components (and their variances) that explain the estimated non-linear manifold,
i.e. matrix $C$ by Algorithm \ref{alg:CFS}, without reconstructing the denoised version of the corrupted data samples. 

\begin{figure}
  \centering 
  \includegraphics[width=.5\linewidth]{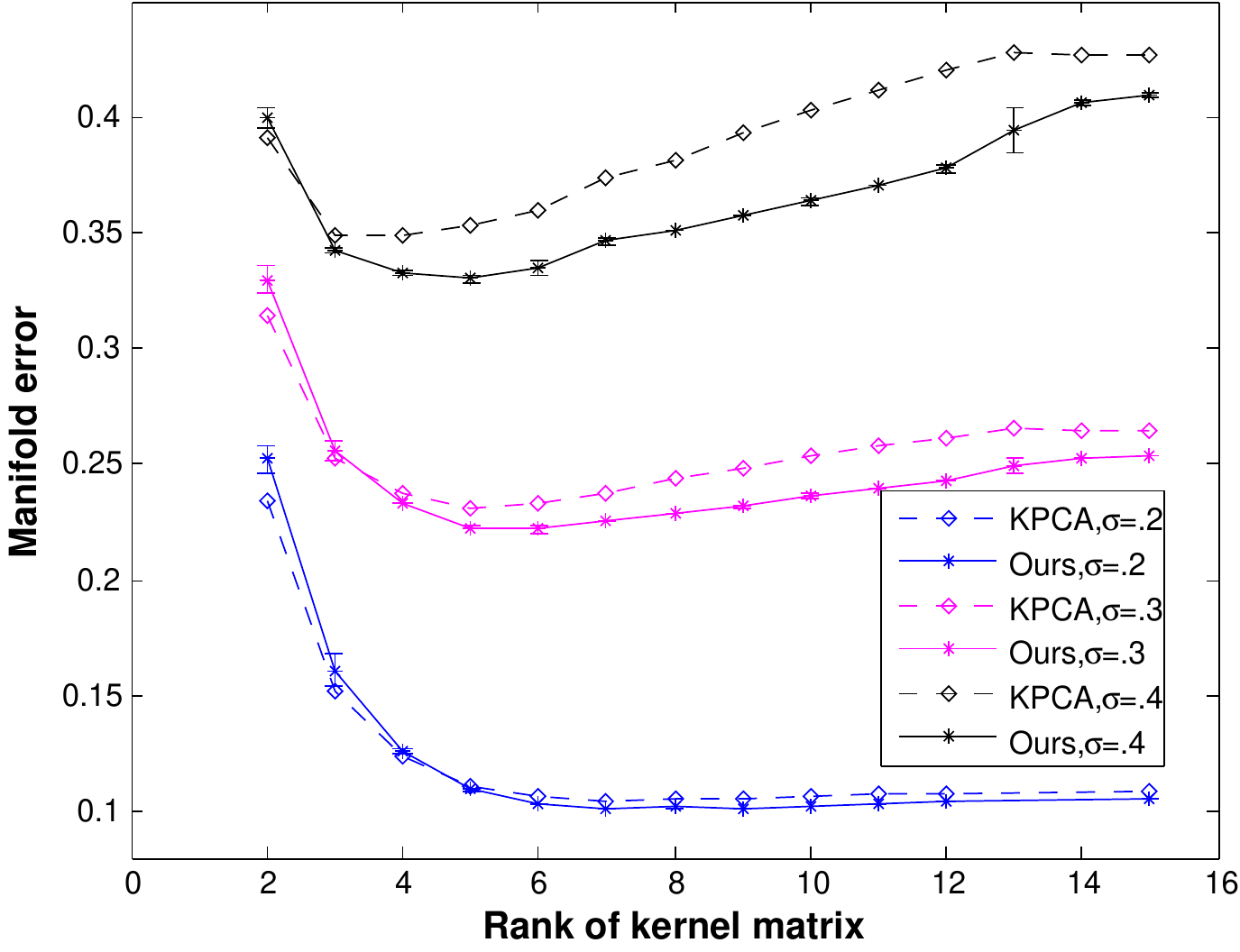}
  \caption{Performance comparison between KPCA and our Robust closed-form solution
   with dimensionality regularization on oil flow dataset with additive Gaussian noise
   of standard deviation $\sigma$. Plots show the normalized kernel matrix errors with different 
   rank of the model. Kernel PCA results are shown in dotted line with diamond while ours are with
   solid line with a star. 
   Bar-plot show the worst and the best errors obtained by our method 
   for a single rank of recovered kernel matrix.\vspace{-3mm}\label{fig:CFS}}
 \end{figure}

Both KPCA and our solution require model selection (choice of rank and $\tau$ respectively) which is beyond the scope of this paper.
Here we resort to evaluate the performance of both methods under different parameters settings. To quantify the
accuracy of the recovered manifold ($C$) we use following criteria:
\begin{itemize}\vspace{-1mm}
 \item Manifold Error : A good manifold should preserve maximum variance of the data --- i.e. it should be able to generate a
 denoised version $K(S_{est}) = C^TC$ of the noisy
 kernel matrix $K(S_\sigma)$. We define the manifold estimation error as $\|K(S_{est}) - K(S_{GT})\|^2_\mathcal{F}$, where $K(S_{GT})$ is the kernel matrix derived using noise free data. 
 Figure \ref{fig:CFS} shows the manifold estimation error for KPCA and our method for different rank and parameter $\tau$ respectively.%
\footnote{Errors from non-noisy kernel matrix can be replaced by cross validating the entries of the kernel matrix for model selection for more realistic experiment.}\vspace{-1mm}
 \item Classification error: The accuracy of a non-linear manifold is often also tested by the nearest neighbor classification accuracy.
 We select the estimated manifold which gives minimum Manifold Error for both the methods and report 1NN classification error (percentage of misclassified example) of the 1000 test points
 by projecting them onto estimated manifolds.
\end{itemize}\vspace{-1mm}

Table \ref{tab:CFS_Accuracy} shows that the proposed method outperforms KPCA to generate less noisy manifold representations with different
ranks and gives better classification results than KPCA for test data. The differences are more significant as the amount of noise increases. 
This simple experiment evaluates our Robust KPCA solution in isolation and indicates that our closed form solution 
itself can be beneficial for the problems where pre-image estimation is not required. 
Note that we have used no loss function $f(W,S)$ in this experiment
however further investigation into more suitable classification loss functions (e.g. \cite{HuangSD12}) should lead to better results.

\begin{table}
\caption{Robust dimensionality reduction accuracy by KPCA versus our closed-form solution on the full oil flow dataset. 
Columns from left to right represent: (1) standard deviation
of the noise in training samples (2-3) Error in the estimated low-dimensional kernel matrix 
by (2) KPCA and (3) our closed-form solution, (4-5) Nearest neighbor classification error of test
data using (4) KPCA and (5) our closed-form solution respectively.
\vspace{-3mm}}\label{tab:CFS_Accuracy}\centering
 \begin{tabular}{|c||c|c||c|c|}
\hline
 	&\multicolumn{2}{c||}{Manifold Error}		& \multicolumn{2}{|c|}{Classification Error}  \\\hline
STD 	 	&KPCA		&Our CFS			& KPCA 			&Our CFS   \\\hline

.2 		&0.1099		&\textbf{0.1068}		&9.60\%			&9.60\%	\\ \hline
.3 		&0.2298		&\textbf{0.2184}		&19.90\%			&\textbf{15.70}\%\\ \hline
.4 		&0.3522		&\textbf{0.3339}		&40.10\%			&\textbf{22.20}\% \\ \hline
\end{tabular}

\end{table}

\subsection{Matrix completion}\label{sec:MatrixComp}
The nuclear norm has been introduced as a low rank prior originally for solving the matrix completion problem. 
Thus, it is natural to evaluate its non-linear extensions on the same task. 
Assuming $W \in R^{m \times n}$ to be the input matrix and $Z$ a binary matrix specifying the 
availability of the observations in $W$, Algorithm \ref{alg:altern} can be used for recovering a complete matrix $S$ with
the following choice of $f(W,Z,S)$:
\begin{equation}
f(W,Z,S)= \| Z \circ( W - S) \|^2_\mathcal{F}
\end{equation}
where $\circ$ represents Hadamard product.

\begin{table}[!t]\caption{Performance comparison on missing data completion on Oil Flow Dataset:
Row 1 shows the amount of missing data 
and subsequent rows show the mean and standard deviation of the error in recovered data matrix over
50 runs on 100 samples of oil flow dataset by: (1) The mean method (also the initialization of other methods)
where the missing entries are replaced by the mean of the known values of the corresponding attributes, (2) 
1-nearest neighbor method in which missing entries are filled by the values of the nearest point, (3) 
PPCA \cite{tipping1999probabilistic}, (4) PKPCA of \cite{sanguinetti2006missing}, (5)RKPCA
\cite{nguyen2009robust} and our method.\vspace{-3mm}}\label{tab:missingData_oilfFlow}
\label{tab:missingData_oilfFlow}\centering
\begin{tabular}{|c||c|c|c|c|}
\hline
p(del) 	&0.05		&0.10 		&0.25 		&0.50  \\\hline
mean   	&13 $\pm$ 4	&28 $\pm$ 4 	&70 $\pm$ 9 	&139 $\pm$ 7 \\ \hline
1-NN 	&5 $\pm$ 3 	&14 $\pm$ 5 	&90 $\pm$ 20 	&NA \\ \hline
PPCA 	&3.7 $\pm$ .6 	&9 $\pm$ 2 	&50 $\pm$ 10 	&140 $\pm$ 30 \\ \hline
PKPCA 	&5 $\pm$ 1 	&12 $\pm$ 3 	&32 $\pm$ 6 	&100 $\pm$ 20 \\ \hline
RKPCA 	&3.2 $\pm$ 1.9 	&8 $\pm$ 4 	&27 $\pm$ 8 	&83 $\pm$ 15  \\\hline

Ours	&\textbf{2.3$\pm$2}
&\textbf{6$\pm$3}		
&\textbf{22$\pm$7}		
&\textbf{70$\pm$11}  \\ \hline
\end{tabular}

\end{table}


To demonstrate the robustness of our algorithm for matrix completion problem,
we choose 100 training samples from the oil flow dataset described in section \ref{sec:CFS} and randomly remove the elements from
the data with varying range of probabilities to test the performance of the proposed algorithm against various baselines.
Following the experimental setup as specified in \cite{sanguinetti2006missing}, we repeat the experiments with 50 different samples of $Z$.
We report the mean and standard deviation of the root mean square reconstruction error for our method with the choice of $\tau = 0.1$, alongside
five different methods in Table \ref{tab:missingData_oilfFlow}.
Our method significantly improves the performance of missing data completion
compared to other robust extensions of KPCA \cite{tipping1999probabilistic,sanguinetti2006missing,nguyen2009robust}, 
for every probability of missing data.

Although we restrict our experiments to least-squares cost functions,
it is vital to restate here that our framework could trivially incorporate robust functions like the $L_1$ norm instead
of the Frobenius norm --- as a robust data term $f(W,Z,S)$ --- to generalize algorithms like Robust PCA \cite{NIPS2009_3704} to their non-linear counterparts.

\subsection{Kernel non-rigid structure from motion}\label{sec:KNRSfM}
Non-rigid structure from motion under orthography is an ill-posed problem where the goal is to
estimate the camera locations and 3D structure of a deformable objects from a collection of 2D images which are labeled with landmark correspondences \cite{bregler2000recovering}.

Assuming $s_i(x_j) \in \R^3$ to be the 3D location of point $x_j$ on the deformable object in the $i^{th}$ image, its orthographic projection $w_i(x_j) \in \R^2$ can be written as $w_{i}(x) = R_i s_i(x_j)$, where $R_i \in \R^{2\times3}$ is a orthographic projection matrix \cite{bregler2000recovering}. 
Notice that as the object deforms, even with
given camera poses, reconstructing the sequence by least-squares reprojection error minimization is an ill-posed problem.
In their seminal work,  \cite{bregler2000recovering} proposed to solve this problem with an additional assumption that the 
reconstructed shapes lie on a low-dimensional linear subspace and can be parameterized as linear combinations of a relatively low number of basis shapes.
NRSfM was then cast as the low-rank factorization problem of estimating these basis shapes and corresponding coefficients.

Recent work, like \cite{dai2014simple,garg2013dense} have shown that the trace norm regularizer can be used as a convex envelope of the low-rank prior to robustly address ill-posed nature of the problem. 
A good solution to NRSfM can be achieved by optimizing:\vspace{-2mm}
\begin{align}
  \min_{S,R} \tau \| S \|_{*} &+ \sum\limits_{i = 1}^F \sum\limits_{j=1}^N Z_i(x_j) \|w_i(x_j) - R_i s_i(x_j)\|_\mathcal{F}^2 \label{eq:NRSfM_Linear}
 \end{align}
where $S$ is the shape matrix whose columns are $3N$ dimensional vectors storing the 3D coordinates $S_i(x_j)$ of the shapes 
and $Z_i(x_j)$ is a binary variable indicating if projection of point $x_j$ is available in the image $i$. 

Assuming the projection matrices to be fixed, this problem is convex and can be exactly solved with standard convex optimization methods. 
Additionally, if the 2D projections $w_i(x_j)$ are noise free, optimizing \eqref{eq:NRSfM_Linear} with very small $\tau$ corresponds to selecting the
the solution --- out of the many solutions --- with (almost) zero projection error, which has minimum trace norm \cite{dai2014simple}. 
Thus henceforth, optimization of \eqref{eq:NRSfM_Linear} 
is referred as the trace norm heuristics (TNH).
We solve this problem with a first order primal-dual variant of the algorithm given in \cite{garg2013dense}, which can handle missing data. The algorithm is detailed 
and compared with other NRSfM methods favorably in the supplementary material.%
\footnote{TNH is used as a strong baseline and has been validated on the full length CMU mocap sequences.
It marginally outperforms \cite{dai2014simple} which is known to be the state of the art  NRSfM approach without missing data.}%

A simple kernel extension of the above optimization problem 
is:
\begin{equation}
 \min_{S,R} \ \ \tau \| \Phi(S) \|_{*} + \underbrace{\sum\limits_{i = 1}^F \sum\limits_{j=1}^N Z_i(x_j) \|w_i(x_j) - R_i s_i(x_j)\|_\mathcal{F}^2}_{f(W,Z,R,S)}  \label{eq:NRSfM_nonlinear}
\end{equation}
where $\Phi(S)$ is the non-linear mapping of $S$ to the feature space using an RBF kernel.

With fixed projection matrices $R$, \eqref{eq:NRSfM_nonlinear} is of the general form \eqref{eq:objectGen}, for which the local optima can be found using Algorithm \ref{alg:altern}.
To solve NRSfM problem with unknown projection matrices, 
we parameterize each $R_i$ with quaternions and alternate between refining the 3D shapes $S$
and projection matrices $R$ using LM. 

\subsubsection{Results on the CMU dataset}
\begin{table}[!t]\caption{3D reconstruction errors for linear and non-linear dimensionality regularization with ground truth camera poses.
Column 1 and 4 gives gives error for TNH while column (2-3) and (5-6) gives the corresponding error for proposed method with different width of RBF kernel.
Row 5 reports the mean error over 4 sequences.\vspace{-3mm}}\label{tab:NRSfM_GTrot}
\centering
\label{tab:NRSfM_GTrot}
\begin{tabular}{|p{.5in}|p{.5in}|p{.5in}|p{.5in}||p{.5in}|p{.5in}|p{.5in}|p{.5in}|p{.5in}|p{.5in}|}
\hline
\multirow{4}{*}{Dataset}  &\multicolumn{3}{c||}{No Missing Data} 			  &\multicolumn{3}{c|}{$50\%$ Missing Data} \\
&\multirow{1}{*}{\textbf{Linear}} & \multicolumn{2}{c||}{\textbf{Non-Linear}} & \multirow{1}{*}{\textbf{Linear}} & \multicolumn{2}{c|}{\textbf{Non-Linear}} \\
			 &	&$d_{max}$  		&$d_{med}$ 			&	        &$d_{max}$  		& $d_{med}$ \\ \hline
Drink		&0.0227		&0.0114			&\textbf{0.0083}		&0.0313		&0.0248			&\textbf{0.0229}\\ \hline
Pickup		&0.0487		&0.0312			&\textbf{0.0279}		&0.0936		&0.0709			&\textbf{0.0658}\\ \hline
Yoga		&0.0344		&\textbf{0.0257}	&0.0276				&0.0828		&\textbf{0.0611}	&0.0612\\ \hline
Stretch		&0.0418		&0.0286			&\textbf{0.0271}		&0.0911		&\textbf{0.0694}	&0.0705\\ \hline
Mean 		&0.0369   	&0.0242			&\textbf{0.0227}		&0.0747		&0.0565			&\textbf{0.0551}\\ \hline
\end{tabular}

\end{table}

We use a sub-sampled version of CMU mocap dataset by selecting every $10^{th}$ frame of the smoothly deforming human body consisting 41 mocap points used in \cite{dai2014simple}.\footnote{
Since our main goal is to validate the usefulness of the proposed non-linear dimensionality regularizer, we opt for a reduced size dataset for more rapid and flexible evaluation.}

In our first set of experiments  we use ground truth camera projection matrices to compare our algorithm against TNH.
The advantage of this setup is that with ground-truth rotation and no noise, 
we can avoid the model selection (finding optimal regularization strength $\tau$) by setting it low enough. 
We run the TNH with $\tau = 10^{-7}$ and use this reconstruction as initialization for Algorithm \ref{alg:altern}. 
For the proposed method, we set $\tau = 10^{-4}$ and use following RBF kernel width selection approach:
\begin{itemize}\vspace{-1mm}
 \item Maximum distance criterion ($d_{max}$): we set the maximum distance in the feature space to be $3\sigma$.
 Thus, the kernel matrix entry corresponding to the shape pairs obtained by TNH with maximum Euclidean distance becomes $e^{-9/2}$. \vspace{-1mm}
 \item Median distance criterion ($d_{med}$): the kernel matrix entry corresponding to the median euclidean distance is set to 0.5.
\end{itemize}\vspace{-1mm}

Following the standard protocol in \cite{dai2014simple,akhter2009nonrigid}, we quantify the reconstruction results with normalized mean 3D errors $e_{3D} = \frac{1}{\sigma F N} \sum_i\sum_j e_{ij}$, 
where $e_{ij}$ is the euclidean distance of a reconstructed point $j$ in frame $i$ from the ground truth,  $\sigma$ is the mean of standard deviation for 3 coordinates for the ground truth 3D structures, and $F,N$ are number of input images and number of points reconstructed.

%

Table \ref{tab:NRSfM_GTrot} shows the results of the TNH and non-linear dimensionality regularization based methods using the experimental setup explained above, both without missing data and after randomly removing 50\% of the image measurements.
Our method consistently beats the TNH baseline and improves the mean reconstruction error by 
$\sim 40\%$ with full data and by $\sim 25\%$ when used with 50\% missing data.
Figure \ref{fig:results_nrsfm} shows qualitative comparison of the obtained 3D reconstruction 
using TNH and proposed non-lienar dimensionality regularization technique for some sample frames from various sequences.
We refer readers to supplementary material for more visualizations.

Table \ref{tab:NRSfM_Optrot} shows the reconstruction performance on a more  realistic experimental setup, with the modification that the camera projection matrices are
initialized with rigid factorization and were refined with the shapes by optimizing \eqref{tab:NRSfM_GTrot}.
The regularization strength $\tau$ was selected for the TNH method by golden section search and parabolic interpolation 
for every test case independently. This ensures the best possible performance for the baseline. 
For our proposed approach $\tau$ was kept to $10^{-4}$ for all sequences for both missing data and full data NRSfM. 
%
This experimental protocol somewhat disadvantages the non-linear method, since its performance can be further improved by a judicious choice of the regularization strength.

However our purpose is primarily to show that the non-linear method adds value 
even without time-consuming per-sequence tuning. 
To that end, note that despite large errors in the camera pose estimations by TNH and $50\%$ missing measurements, the proposed method shows significant ($\sim 10\%$) improvements in terms of 
reconstruction errors, proving our broader claims that non-linear representations are better suited for modeling real data,
and that our robust dimensionality regularizer can improve inference for ill-posed problems.

\begin{table}[!t]\caption{3D reconstruction errors for linear and non-linear dimensionality regularization with noisy camera pose initialization from rigid factorization and refined in alternation with shape.
The format is same as Table \ref{tab:NRSfM_GTrot}.\vspace{-3mm}}\label{tab:NRSfM_Optrot}\centering
\label{tab:NRSfM_Optrot}
\begin{small}
\begin{tabular}{|p{.5in}|p{.5in}|p{.5in}|p{.5in}||p{.5in}|p{.5in}|p{.5in}|p{.5in}|p{.5in}|p{.5in}|}
\hline
\multirow{4}{*}{Dataset} &\multicolumn{3}{c||}{No Missing Data} 			  &\multicolumn{3}{c|}{$50\%$ Missing Data} \\
			 &\multirow{1}{*}{\textbf{Linear}} & \multicolumn{2}{c||}{\textbf{Non-Linear}} & \multirow{1}{*}{\textbf{Linear}} & \multicolumn{2}{c|}{\textbf{Non-Linear}}\\
			 &$\scriptsize{\tau=\tau^*}$		&\multicolumn{2}{c||}{\scriptsize{$\tau = 10^{-4}$}}  	  		&$\scriptsize{\tau=\tau^*}$		&\multicolumn{2}{c|}{\scriptsize{$\tau = 10^{-4}$}}\\
			 &		&$d_{max}$	  	&$d_{med}$	 	&			&$d_{max}$  		& $d_{med}$ \\ \hline
Drink			&0.0947		& 0.0926		&\textbf{0.0906}			&0.0957			&0.0942			&\textbf{0.0937}	\\ \hline
Pickup			&0.1282		&0.1071			&\textbf{0.1059}	&0.1598			&0.1354			&\textbf{0.1339}	\\ \hline
Yoga			&0.2912		&0.2683			&\textbf{0.2639}	&0.2821			&\textbf{0.2455}	&\textbf{0.2457}	\\ \hline
Stretch			&0.1094		&0.1043			&\textbf{0.1031}	&\textbf{0.1398}	&0.1459			&0.1484	\\ \hline
Mean 			&0.1559		&0.1430			&\textbf{0.1409}	&0.1694	   		&\textbf{0.1552}	&\textbf{0.1554} 	\\ \hline
\end{tabular}
\end{small}
\end{table}

As suggested by \cite{dai2014simple}, robust camera pose initialization is beneficial for the structure estimation. We have used rigid factorization
for initializing camera poses here but this can be trivially changed.
We hope that further improvements can be made by choosing better kernel functions, 
with cross validation based model selection (value of $\tau$) and with a more appropriate tuning of kernel width.
Selecting a suitable kernel and its parameters is crucial for success of kernelized algorithms. It becomes more challenging 
when no training data is available. We hope to explore other kernel functions and parameter selection criteria in our future work.  
%


We would also like to contrast our work with \cite{gotardo2011kernel}, which is the only work we are aware of
where non-linear dimensionality reduction is attempted for NRSfM. 
While estimating the shapes lying on a two dimensional non-linear manifold, \cite{gotardo2011kernel} additionally assumes smooth 3D trajectories 
(parametrized with a low frequency DCT basis) and a pre-defined hard linear rank constraint on 3D shapes.
The method relies on sparse approximation of the kernel matrix as a proxy for dimensionality reduction.
The reported results were hard to replicate under our experimental setup for a fair comparison due to non-smooth deformations.
However, in contrast to \cite{gotardo2011kernel}, our algorithm is applicable in a more general setup, can be modified to incorporate smoothness priors  
and robust data terms but more importantly, is flexible to integrate with a wide range of energy minimization formulations leading to a larger applicability beyond NRSfM.

\section{Conclusion}
In this paper we have introduced a novel non-linear dimensionality regularizer
which can be incorporated into an energy minimization framework, while solving an inverse problem.
The proposed algorithm for penalizing the rank of the data in the feature space has been shown to be robust
to noise and missing observations. 
We have picked NRSfM as an application to substantiate our arguments and have shown that despite missing data and model noise (such as erroneous camera poses)
our algorithm significantly outperforms state-of-the-art linear counterparts.

Although our algorithm currently uses slow solvers such as the penalty method and is not directly scalable to very large problems like dense non-rigid reconstruction,
we are actively considering alternatives to overcome these limitations. 
An extension to estimate pre-images with a problem-specific loss function is possible, and this will be useful for online inference with pre-learned low-dimensional manifolds. 

Given the success of non-linear dimensionality reduction in modeling real data and overwhelming use of the linear 
dimensionality regularizers in solving real world problems, we expect that proposed non-linear dimensionality regularizer will  be applicable to a wide variety of unsupervised inference problems: recommender systems; 3D reconstruction; denoising; shape prior based object segmentation; and tracking are all possible applications.

\appendix
\section{Proof of Theorem 3.1}\label{Appendix:Proof}
 \begin{proof}\begin{small}
 We will prove theorem  \ref{th:CFS} by first establishing a lower bound for \eqref{eq1a} and subsequently showing that this lower bound is obtained at $L^*$ given by \eqref{eq1c}.
 The rotational invariance of the entering norms allows us to write \eqref{eq1a}
 as:%
\begin{align} 
 \min_{
 \arraj{
 \Gamma \in \D^n,\\
  W^T W=I
 }
 } \frac{\rho}{2} \normF{\Sigma- W \Gamma^2 W^T} + 
 \tau \normnuc{\Gamma}. 
 \label{eq2_sup}
 \end{align}
 Expanding \eqref{eq2_sup} we obtain
 \begin{align}
 &\frac{\rho}{2} \min_{  \Gamma , W } 
  \trace{\Sigma^2} - 
 2 \trace{\Sigma W \Gamma^2 W^T} + \trace{\Gamma^4}  +  
 \frac{2\tau}{\rho} \sum_{i=1}^n \gamma_i  \\
 &=\frac{\rho}{2} \min_{ \Gamma , W} 
 \sum_{i=1}^n \left(\sigma_i^2 + \gamma_i^4 + \frac{2\tau}{\rho} \gamma_i \right) 
 -2 \sum_{i=1}^n \sum_{j=1}^n  w_{ij}^2  \gamma_j^2 \sigma_i \label{eq3a_sup}\\
 %
 &\geq \frac{\rho}{2} \min_{ \arraj{ \Gamma }} 
 \sum_{i=1}^n \left(\sigma_i^2 -2 \gamma_i^2 \sigma_i+ \gamma_i^4 + \frac{2\tau}{\rho} 
 \gamma_i \right) \label{inqual_sup}\\
 &=\frac{\rho}{2} \sum_{i=1}^n \min_{\gamma_i \geq 0} 
 \left(  
 \sigma_i^2 -2 \gamma_i^2 \sigma_i+ \gamma_i^4 + \frac{2\tau}{\rho} \gamma_i 
 \right) 
 \label{eq4d_sup}
 %
 \end{align} 
 The inequality in \eqref{inqual_sup} follows directly by applying H\"older's inequality to \eqref{eq3a_sup} and 
 using the property that the column vectors $w_i$ are unitary. 
 
 Next, with $L=\Gamma U^T$ in \eqref{eq1a} we have
 \begin{align}
 \frac{\rho}{2} \normF{A-L^T L} + &\tau \normnuc{L} =  
 \frac{\rho}{2} \normF{\Sigma-\Gamma^2} + \tau \normnuc{\Gamma} \nonumber\\
 &=\sum_{i=1}^n \left(  \frac{\rho}{2} (\sigma_i- \gamma_i^{2} )^2 + \tau \gamma_i  \right).
 \end{align}
 Finally, since the subproblems in \eqref{eq4d_sup} are separable in $\gamma_i$, its minimizer 
 must be KKT-points of the individual subproblems. As the constraints are simple non-negativity 
 constraints, these KKT points are either (positive) stationary points of the objective functions or $0$. 
 It is simple to verify that the stationary points are given by the roots of the cubic function 
 $p_{\sigma_i,\tau / 2\rho}$. Hence it follows that there exists a $\gamma^*_i$ such that
 \begin{align}
 \frac{\rho}{2} \left(  \sigma_i^2 -2 \gamma_i^2 \sigma_i+ \gamma_i^4 + \frac{2\tau}{\rho} \gamma_i \right) 
 \geq
 \frac{\rho}{2} (\sigma_i- \gamma_i^{*2} )^2 + \tau \gamma_i^{*},
 \label{eq5}
 \end{align}
 $\forall \gamma_i \geq 0$, which completes the proof.  
 \end{small}
  \end{proof}

\clearpage

\bibliographystyle{splncs03}
\bibliography{egbib_1}
\end{document}

%% file: Macros.tex
\def \R{{\mathbb{R}}}

\def \D{{\mathcal{D}}}

\def \S{{\mathcal{S}}}

\DeclareMathOperator*{\tr}{tr}
\newcommand{\trace}[1]{ \tr \left(  #1 \right)}

\newcommand{\arraj}[1]{  \begin{smallmatrix} #1 \end{smallmatrix} }

\newcommand {\normF}[1] { || #1 ||^2_\mathcal{F}}
\newcommand {\normnuc}[1] { || #1 ||_*}

\newcolumntype{L}[1]{>{\raggedright\let\newline\\\arraybackslash\hspace{0pt}}m{#1}}
\newcolumntype{C}[1]{>{\centering\let\newline\\\arraybackslash\hspace{0pt}}m{#1}}
\newcolumntype{R}[1]{>{\raggedleft\let\newline\\\arraybackslash\hspace{0pt}}m{#1}}